\documentclass[runningheads]{llncs}
\usepackage{graphicx}
\usepackage{amsmath,amssymb} 
\usepackage{color}
\usepackage{ifthen}
\usepackage{bigstrut}
\usepackage[pagebackref=true,breaklinks=true,letterpaper=true,colorlinks,bookmarks=false]{hyperref}

\definecolor{frenchblue}{rgb}{0.0, 0.45, 0.73}
\definecolor{gray}{rgb}{0.5,0.5,0.5} 
\definecolor{green}{rgb}{0, 0.4, 0} 
\definecolor{orange}{rgb}{1, 0.5, 0} 	
\definecolor{mahogany}{rgb}{0.75, 0.25, 0.0}
\definecolor{purple}{rgb}{0.6, 0, 0.6}
\definecolor{darkgreen}{rgb}{0, 0.4, 0.4} 
\definecolor{blue}{rgb}{0.0, 0.45, 0.73}
\definecolor{aaaa}{rgb}{0.55, 0.1, 0.7}
\newboolean{revising}
\setboolean{revising}{false}
\ifthenelse{\boolean{revising}}
{
	\newcommand{\ignore}[1]{}

	\newcommand{\horatio}[1]{\textcolor{blue}{#1}}

	\newcommand{\hu}[1]{\textcolor{orange}{#1}}

} {
	\newcommand{\ignore}[1]{}

	\newcommand{\horatio}[1]{#1}

	\newcommand{\hu}[1]{#1}

}

\begin{document}

\title{Self-Supervised Learning of Depth and Camera Motion from 360$^\circ$ Videos} 
\titlerunning{Self-Supervised 360$^\circ$ Depth}

\author{
    $^1$Fu-En Wang$^*$  \and
    $^1$Hou-Ning Hu$^*$  \and
    $^1$Hsien-Tzu Cheng$^*$  \and
    $^1$Juan-Ting Lin  \and
    $^2$Shang-Ta Yang  \and
    $^1$Meng-Li Shih  \and
    $^2$Hung-Kuo Chu  \and
    $^1$Min Sun 
}

\authorrunning{Wang, Hu and Cheng et al.} 

\institute{
    $^1$Dept. of Electrical Engineering, National Tsing Hua University, Taiwan \\
    $^2$Dept. of Computer Science, National Tsing Hua University, Taiwan \\
    \{fulton84717, eborboihuc, hsientzucheng, brade31919, sundadenny, shihsml, hkchu\}@gapp.nthu.edu.tw, sunmin@ee.nthu.edu.tw \\
    $^*$The authors contribute equally to this paper.
}

\maketitle

\begin{abstract}
As 360$^{\circ}$ cameras become prevalent in many autonomous systems (e.g., self-driving cars and drones), efficient 360$^{\circ}$ perception becomes more and more important. 
We propose a novel self-supervised learning approach for predicting the omnidirectional depth and camera motion from a 360$^{\circ}$ video.
In particular, starting from the SfMLearner, which is designed for cameras with normal field-of-view, we introduce three key features to process 360$^{\circ}$ images efficiently.
Firstly, we convert each image from equirectangular projection to cubic projection in order to avoid image distortion. In each network layer, we use Cube Padding (CP), which pads intermediate features from adjacent faces, to avoid image boundaries.
Secondly, we propose a novel ``spherical" photometric consistency constraint on the whole viewing sphere. In this way, no pixel will be projected outside the image boundary which typically happens in images with normal field-of-view.
Finally, rather than naively estimating six independent camera motions (i.e., naively applying SfM-Learner to each face on a cube), we propose a novel camera pose consistency loss to ensure the estimated camera motions reaching consensus.
To train and evaluate our approach, we collect a new PanoSUNCG dataset containing a large amount of 360$^{\circ}$ videos with groundtruth depth and camera motion. Our approach achieves state-of-the-art depth prediction and camera motion estimation on PanoSUNCG with faster inference speed comparing to equirectangular. In real-world indoor videos, our approach can also achieve qualitatively reasonable depth prediction by acquiring model pre-trained on PanoSUNCG.

\end{abstract}
\section{Introduction}\label{sec.Intro}
Thanks to the emergence of Virtual Reality (VR) applications, 360$^{\circ}$ cameras have become very popular. Nowadays, one can easily find a few 360$^{\circ}$ cameras supporting real-time streaming with 4K resolution and 30 frames-per-second (fps) at a consumer price.
As a result, watching 360$^{\circ}$ photos and videos is becoming a common experience on sites like YouTube and Facebook. 
For humans, they enjoy the immersive experience in 360$^{\circ}$ videos as they can freely move their viewing angle.
However, we argue that the ability to capture all surrounding at once in high-resolution and high frame-rate is immersively more critical for autonomous systems than for entertaining humans.

All autonomous systems need to perceive the surrounding in order to act in the world safely. This includes self-driving cars, indoor/outdoor robots, drones, and even underwater robots.
Using a single traditional camera, the autonomous system will need to turn its viewing direction many times in order to explore the environment. To avoid such inefficiency, high-end autonomous systems are typically equipped with multiple cameras. However, this introduces extra cost and technical challenge to maintain a stable and well calibrated multiple camera system.
In this case, modern 360$^{\circ}$ cameras are a great alternative since they are well-calibrated, low-cost, and supporting real-time streaming with high-resolution. Hence, we believe in the future, 360$^{\circ}$ cameras will be widely adopted in all kinds of autonomous systems.

\begin{figure}[t!]
\begin{center}
\includegraphics[width=1\linewidth]{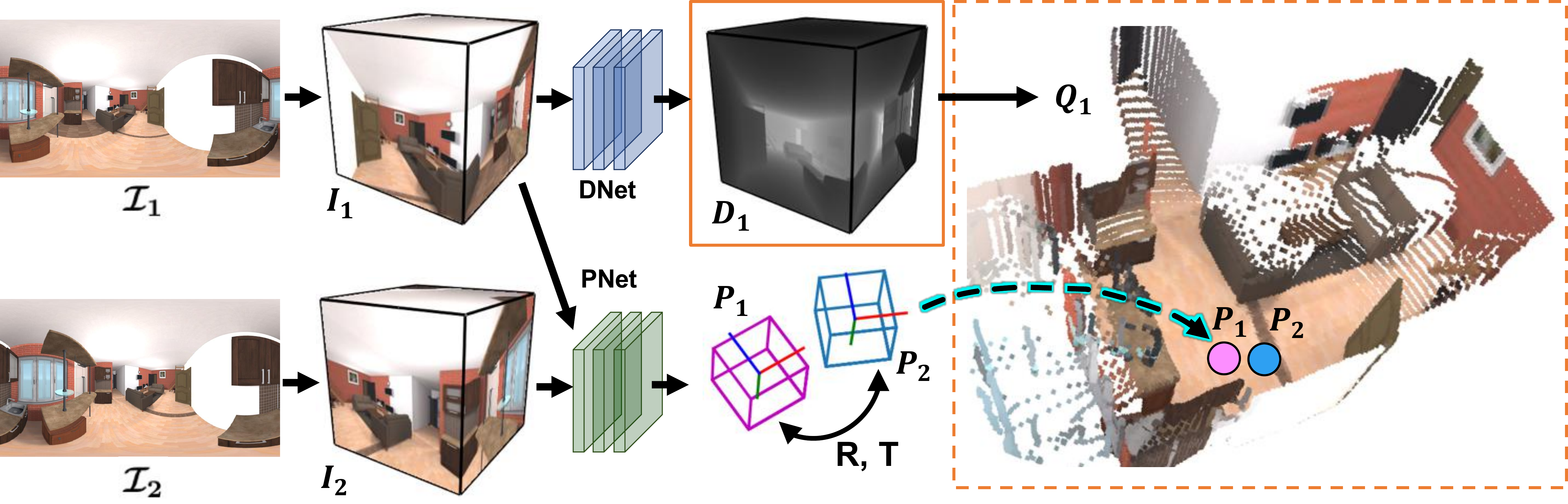}
\end{center}
\caption{
Overview of 360$^{\circ}$ depth and camera motion estimation. 
In training phase, we use two 360$^{\circ}$ images ($\mathcal{I}_1$ and $\mathcal{I}_2$) as our input. 
We first project them into cubemaps ($I_1$ and $I_2$) which consist of 6 faces with 90$^{\circ}$ FoV. 
The DispNet (DNet) estimates depth of $I_1$ and PoseNet (PNet) predicts explainability map and camera motion (i.e., rotation $R$ and translation $T$) between $I_1$ and $I_2$.
With depth map $D_1$, we reproject our image $I_1$ to 3D point cloud $Q_1$.
To update our network, depth and camera motion are used to calculate spherical photometric loss (Sec.~\ref{sec.losses}) as our self-supervised objective.}
\label{fig.overview}
\end{figure}

A few recent methods have proposed to tackle different perception tasks on 360$^\circ$ videos.
Caruso et al.~\cite{caruso2015large} propose a robust SLAM algorithm leveraging the wide FoV in 360$^\circ$ videos. Su and Grauman~\cite{SuNIPS} propose to learn spherical convolution for fast and robust object detection in 360$^\circ$ images. Huang et al.~\cite{1803.04228} propose to recognize place robustly given 360$^\circ$ images.
To the best of our knowledge, predicting depth from a monocular 360$^\circ$ camera has not been well studied. 
We argue that perceiving depth is one of the most important perception ability for autonomous systems since stereo cameras or lidars is the must-have equipment on self-driving cars.

In this paper, we propose a novel deep-learning based model to efficiently predict the depth of each panoramic frame and camera motion between two consecutive frames in a 360$^\circ$ video. 
We propose to train our method under different levels of self-supervision: (1) just raw 360$^\circ$ video, (2) noisy ground truth camera motion and 360$^\circ$ video, and (3) ground truth camera motion and 360$^\circ$ video. Intuitively, higher depth prediction accuracy can be achieved with a higher level of supervision. In practice, noisy ground truth camera motion can be measured by IMU. Our idea of self-supervision is inspired by SfMLearner~\cite{zhou2017unsupervised}. However, since it is designed for traditional cameras with normal field-of-view (NFoV), we introduce several key features to incorporate the unique properties in 360$^\circ$ video. 

\noindent In this paper, our contributions are:

\noindent\textbf{Spherical photo-consistency loss.}
We propose to compute photo-consistency loss on the whole viewing sphere (Fig.~\ref{fig.3d_losses}). In this way, no pixel will be projected outside the image boundary which typically happens in images with normal field-of-view.

\noindent\textbf{Robust motion estimate.}
Rather than naively estimating six independent camera motions (i.e., naively applying SfMLearner to each face on the cube), we propose a novel pose consistency loss to ensure the estimated camera motions reaching consensus (Fig.~\ref{fig.3d_losses}).

\noindent\textbf{PanoSUNCG dataset.}
To train and evaluate our approach, we collect a new PanoSUNCG dataset containing a large amount of 360$^{\circ}$ videos with ground truth depth and camera motion.

We compare our method with variants of our method not including all three key features. Our full method achieves state-of-the-art depth prediction and camera motion estimation on PanoSUNCG with faster inference speed comparing to equirectangular. Once pre-trained on PanoSUNCG, our method can easily be fine-tuned on real-world 360$^{\circ}$ videos and obtain reasonably good qualitative results.

\section{Related work}\label{sec.RW}

\noindent\textbf{Single view depth estimation.}
In navigation and self-driving application, an accurate depth map is usually required to help agent to make a decision. In general, depth can come from \horatio{LiDAR} or stereo images. 
As we know, \horatio{LiDAR} provides us accurate depth. However, the high cost of such sensor makes these application challenging to productize. 
To reduce the cost, using a stereo camera is also common to get depth. 
By this way, the quality of depth will be sensitive to the calibration, which means a small error in calibration will reduce the depth accuracy. 
As deep neural network prospers in recent years, some researches are working on estimating depth map only from a single RGB image. Laina et al.\cite{laina2016deeper} uses fully convolutional architecture with feature map upsampling to improve the regression of prediction and ground truth. 
With the method in \cite{laina2016deeper}, \cite{tateno2017cnn} uses it as depth initialization and combines it with LSD-SLAM \cite{engel2014lsd} to much improve the quality of reconstruction.
With the photometric warping loss, the deep neural network can unsupervisedly learn single depth prediction by using image pairs from unstructured videos.
\cite{vijayanarasimhan2017sfm} unsupervisedly split the scene into several rigid body motion and masks, which is proposed in \cite{byravan2017se3}, and use motion to transform point cloud coming from depth prediction. Zhou et al.\cite{zhou2017unsupervised} uses a multi-scale method to predict single view depth and relative pose from unstructured image pairs. 
For each scale, they apply warping and smoothing loss to each of them. 
They also propose explainability masks to weight sum photometric error to prevent occlusion and moving object problems. There are also some recent works improve the unsupervised method by adding differentiable 3D losses \cite{Mahjourian_2018_CVPR}, replacing pose prediction by Direct Visual Odometry (DVO)\cite{Wang_2018_CVPR}. 
However, all the works above cannot be directly applied to 360$^{\circ}$ images due to the different geometry of spherical projection.

\noindent\textbf{360$^\circ$ perception.}
Omni-directional cameras has an emerging potential in a variety of applications. H{\"a}ne et al.\cite{3DAutonomous} proposed an autonomous driving system with 360$^\circ$ FoV input. A series of works address semantic saliency\cite{lai2017semantic}, focus preference\cite{lin2017tell} and NFoV viewing assistant\cite{hu2017deep360pilot,su2016activity,su2017videography} in 360$^\circ$ panoramic videos. Current methods deal with an equirectangular image or further process them into NFoV crops, while our system process a cubemap which can reach more than 200\% FPS (Fig.~\ref{fig.speed}) in high resolution.
Surrounding scene recorded into a single image frame gives a great advantage to autonomous systems since 360$^\circ$ view ensures full coverage of every detail all around.

With the entire field of view, 360$^\circ$ camera is more robust to rotation and translation movements compared to normal perspective camera \cite{7832307}.
There are some recent methods aim to tackle 360$^\circ$ perception tasks. 
Su and Grauman~\cite{SuNIPS} propose spherical convolution for fast object detection in 360$^\circ$ images. 
Huang et al.~\cite{1803.04228} propose a new system in 360$^\circ$ images that able to recognize place robustly.  Cheng et al.~\cite{Cheng_2018_CVPR} found using cubemap 
representation can help network to predict 360$^\circ$ saliency map very well.
Although a significant amount of information carried by a 360$^\circ$ view, it can hardly be used in a perspective camera system without being explicitly designed. 
Therefore, several works have been proposed in solving 3D reconstruction problems~\cite{im2016all,6942637}, such as SLAM~\cite{caruso2015large}, Structure from Motion~\cite{7707455,Chang-2000-8057,6130266}, camera motion estimation~\cite{4409198}, etc.

However, none of the works above adapts learning-based method to directly mitigate the re-projection problem raised by the nature of 360$^\circ$ FoV. We thus focus on solving the problem regarding speed and robustness by considering a whole panoramic view in one forward.
\section{Our approach}\label{sec.Tech}

In this section, the method that jointly estimates depth and motion from 360$^{\circ}$ videos is introduced. Note that spherical photometric constraints are proposed so as to adapt the model input from NFoV videos to 360$^{\circ}$ videos.
We describe the important components in our method including spherical projection (Sec.~\ref{sec.proj}), 360$^{\circ}$ depth and camera motion estimation (Sec. ~\ref{sec.depth_pose}), self-consistency constraints for self-supervision learning (Sec.~\ref{sec.losses}), and 2D image-based constraints (Sec.~\ref{sec.2dlosses}). Before that, we first introduce the notations used in our formulation in Sec.~\ref{sec.notation}.

\subsection{Preliminaries and Notations}\label{sec.notation}

Given a cubemap, the set of 6 faces---Back, Down, Front, Left, Right, and Up---are denoted as $\mathcal{F} = \{B, D, F, L, R, U\}$, respectively. 
In general, we define 2D equirectangular map as $\mathcal{M}$, and the cubemaps $M = \{M^f|f \in \mathcal{F}\}$, where $M^f$ corresponds to the $f$ face.
We define a function $\Phi $ to transform equirectangular map to cubemap by ${M} = \Phi (\mathcal{M})$. Note that $\mathcal{M}$ represents an image, a feature map, or a depth map when we consider the input, the intermediate feature maps, or output of the network, respectively.

Given a 360$^{\circ}$ video sequence with $T$ frames, we define the equirectangular frames $\mathcal{I} = \{ \mathcal{I}_i| 1 \leq i \leq T\}$, the corresponding cubemap frames $I = \{ I_i=\Phi(\mathcal{I}_i) | 1 \leq i \leq T \}$, the depth map $D = \{D_i|1 \leq i \leq T\}$ and the relative camera pose $P = \{P_{ij}|1 \leq i,j \leq T \}$ where $P_{ij}$ denotes the relative camera pose from $I_i$ to $I_j$. Besides, each relative pose $P_{ij}$ consists of rotation $R_{ij}$ and translation $T_{ij}$ so we can also represent $P_{ij}$ as $(R_{ij}, ~T_{ij})$.

After the depth $D_i$ is attained, it can be projected into a 3D point cloud $Q_i = \Psi(D_i)$.
To further use photometric consistency as our training loss, we define a warping function $\hat{I}_i  = \xi(Q_i, I_j)$ to sample on $I_j$ by the point cloud of $I_i$. The details of $\Psi$ and $\xi$ will be explained in
Sec.~\ref{sec.proj}.

\subsection{Spherical Projection}\label{sec.proj}
\begin{figure}[t]
	\centering
    \includegraphics[width=0.6\textwidth]{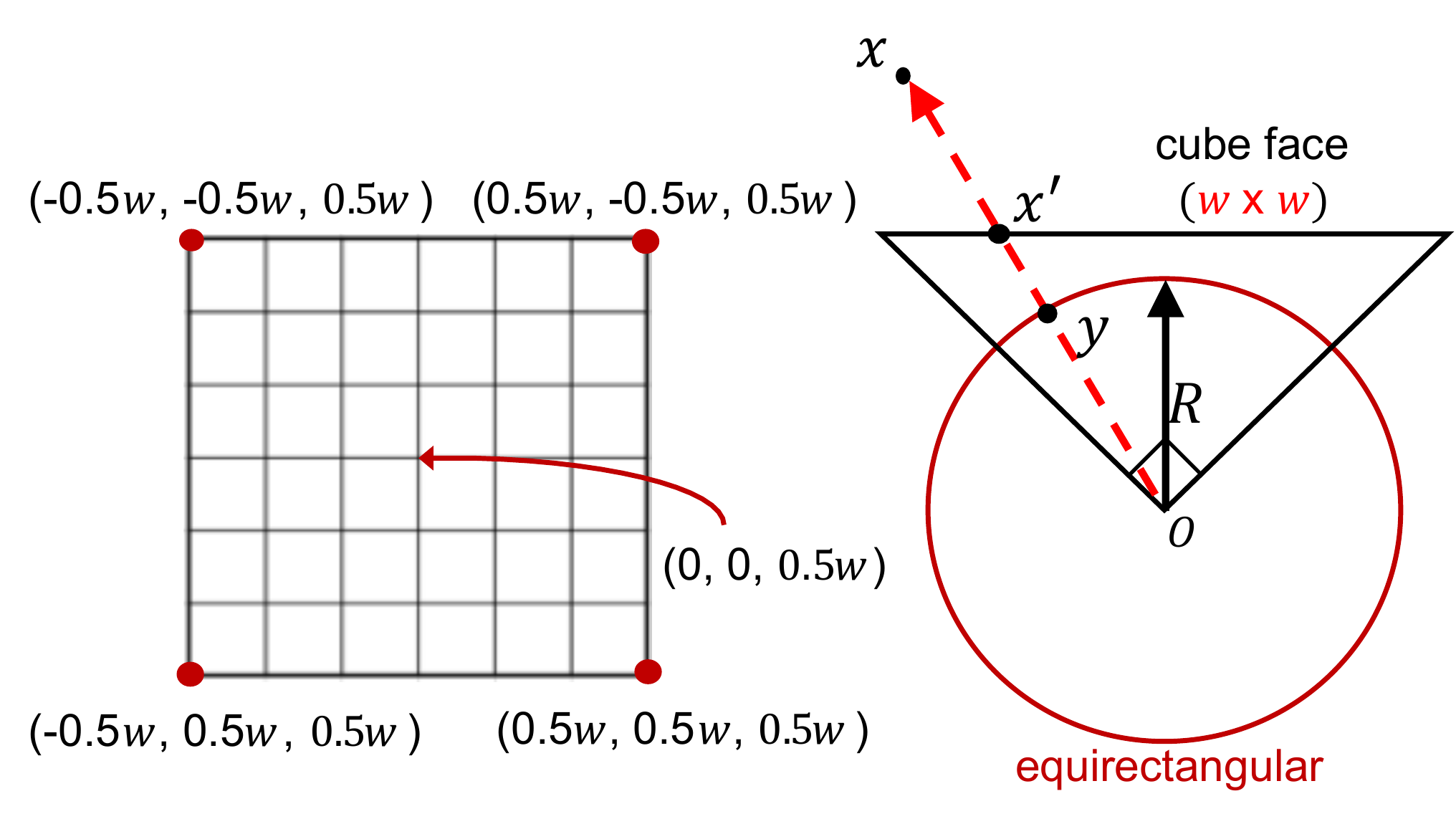}
    \caption{Spherical projection}
    \label{fig.proj}
\end{figure}
For a perfect 360$^\circ$ camera, we can treat its imaging plane as a sphere as shown in Fig.~\ref{fig.proj} (Right). So any point $x$ in world coordinate can be projected onto this sphere as the new point $y$ and this is what we call spherical projection. If we want to reproject points on the sphere onto a cube face, we can assume a new pinhole camera with 90$^\circ$ FoV and its dimension is $w \times w$. Because we know its FoV is 90$^o$, we can obtain that its focal length is $\frac{w}{2}$. For a pinhole camera, we can assume the 3D coordinate of each pixel in imaging plane is divided equally. So we can get a 3D grid $G$ of this imaging plane as in Fig.~\ref{fig.proj} (Left). Notice that all the z value will be $0.5w$.

Now we use $E(\theta),~\theta=(\theta_x, \theta_y, \theta_z)$ to represent the transformation from 
euler angle $(\theta_x, \theta_y, \theta_z)$ to rotation matrix. For the six faces of a cubemap, the relative
rotation respecting to the front face is 

$\{(0,\pi,0), (-0.5\pi,0,0), (0,0,0), (0,-0.5\pi,0), (0,0.5\pi,0), (0.5\pi,0,0)\}$ 
in the order of $\{Back, Down, Front, Left, Right, Up\}$ respectively. Now if we want to get 
right face, we need calculate the grid of right face by
$$
	G^R = E(0,0.5\pi,0) \cdot G~.
$$

With the grid of the right face $G^R$, the correspondence between pixels on the cubemap and equirectangular image can be established as the following equation:
\begin{align}
	X &= \frac{arctan_2{(\frac{x}{z}})}{\pi}~,
    \label{eq:warp-x} \\
    Y &= \frac{arcsin{(\frac{y}{\sqrt{x^2+y^2+z^2}}})}{0.5\pi}~. 
    \label{eq:warp-y}
\end{align}
where $X$ and $Y$ are normalized pixel location on the 
equirectangular image and $(x, y, z)$ is 3D point in the grid $G^R$. With the correspondence between pixel in cube face
and equirectangular image, we can simply use inverse warping interpolation to transform 360$^\circ$ image to any 
cube faces. In this paper, we refer $\Phi$ as the function to project an equirectangular image to a cubemap with 6 faces.

In Sec.~\ref{sec.notation}, we use the function $\Psi$ to convert depth map of 6 cube faces to point cloud.
The transformation is to scale up the unit grid $G^i$, where $i \in \mathcal{F}$, according to the depth.
Take the right face for example, if we want to convert a pixel which location is $(X, Y)$ and depth is $d$,
we just need to normalize the corresponding grid location $(x, y, z)$ in $G^R$ and scale up to the same length as the 
depth. For the warping function, we use $\hat{I_i}=\xi(P_{ij} \cdot Q_i, I_j)$ to indicate the warping between $I_i$ and $I_j$, the term $P_{ij} \cdot Q_i$ simply means the new point cloud transformed by the relative camera pose. And $\xi$ means  projecting point cloud to pixel coordinate of another sphere by Eq.~\ref{eq:warp-x} 
and \ref{eq:warp-y} and then applying interpolation.

\subsection{Model for Depth and Camera Motion Estimation}\label{sec.depth_pose}

Similar to \cite{zhou2017unsupervised}, we adopt two networks for depth and camera motion estimation individually and join them to train together with spherical photometric and pose constraints. 
For DispNet, we utilize architecture refer to \cite{mayer_dispnet} that takes a single RGB image to predict the corresponding depth map. 
For PoseNet, refer to \cite{zhou2017unsupervised}, our network takes the concatenation of target view with all the source views as input, then output relative poses between target and source. 
Another output is explainability mask (elaborated in Sec.~\ref{sec.exp_mask}) which can improve depth network accuracy by masking out ambiguity areas like occlusion and non-static patches. 
In our entire networks, every layer with 2D operations, such as convolution, pooling, etc., are incorporated with Cube Padding module as proposed by 
\cite{Cheng_2018_CVPR} to connect the intermediate features at adjacent faces and make the six faces spatially continuous.

\subsection{Self Consistency Constraints}\label{sec.losses}
To get supervision from the 3D environment and spherical camera geometry, here we introduce two self-consistency constraints: photometric consistency and camera pose consistency constraints for self-learning. 

\begin{figure}[h!]
\begin{center}
\includegraphics[width=1\linewidth]{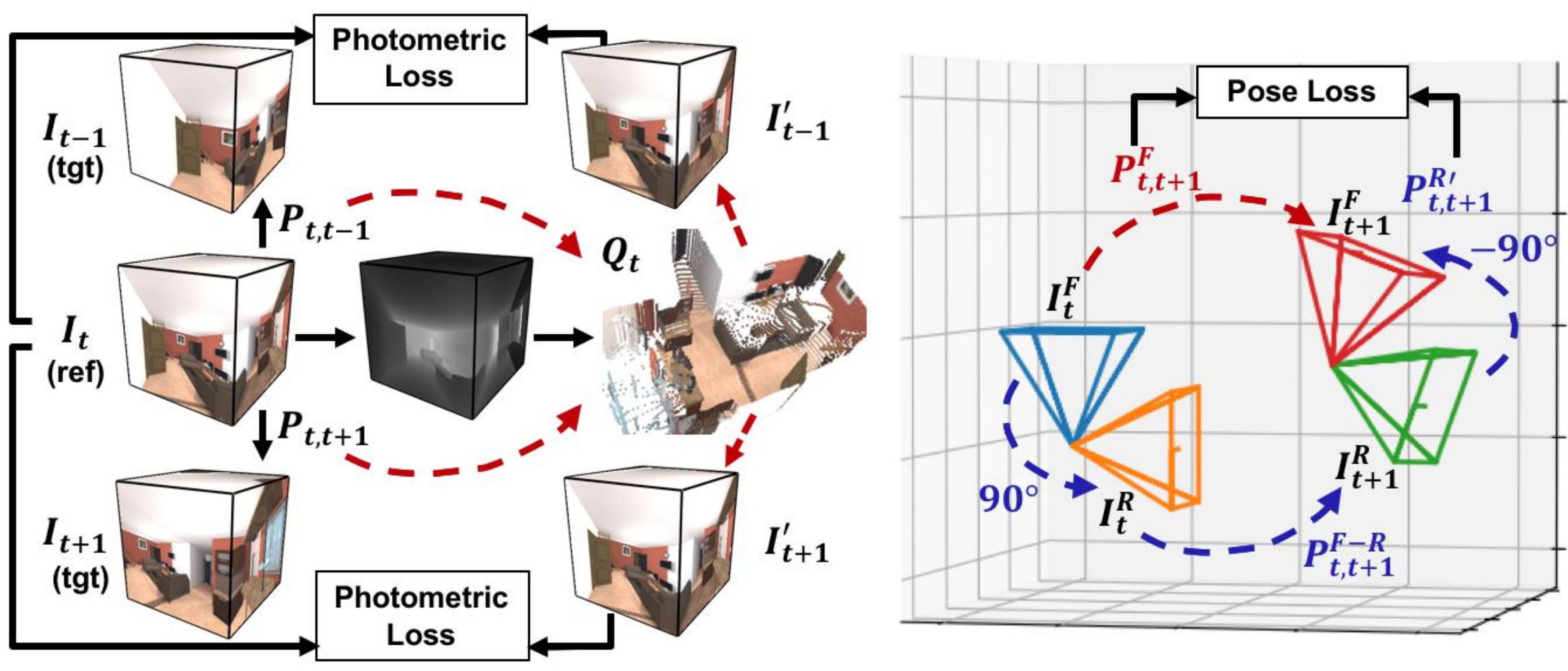}
\end{center}
\caption{
Spherical photometric constraints. Left panel: photometric consistency loss. Right panel: pose consistency loss. In the left panel, $I_{t-1}$ and $I_{t+1}$ are target (tgt) frames of the reference (ref) $I_{t}$ with relative pose $P_{t,t-1}, P_{t,t+1}$. Firstly, $Q_t$ is projected by $I_t$ through $D_t$. $I_{t-1}^\prime$ and $I_{t+1}^\prime$ are generated by transforming $Q_t$ via ${P_{t,t-1}}$, $P_{t,t+1}$ and sampling from $I_{t-1}$, $I_{t+1}$, respectively. Then, we compute the photometric loss of reprojected $I_{t-1}^\prime$ and $I_{t+1}^\prime$ with $I_t$ .
On the right side, the original relative pose between $I^F_{t}$ and $I^F_{t+1}$ is denoted as $P^F_{t,t+1}$. Regarding the 90$^{\circ}$ rotation between cube faces, we can get $P^{R\prime}_{t,t+1}$ from the right face $P^R_{t,t+1}$ through $P^{F-R}_{t,t+1}$. Eventually we can compute pose consistency loss between $P^{R\prime}_{t,t+1}$ and $P^F_{t,t+1}$.
}
\label{fig.3d_losses}
\end{figure}

\noindent\textbf{Photometric consistency.}
By synthesizing a target view, given the relative pose, depth, and visibility in nearby views, the difference between the warped image and target image can be treated as a supervision cue~\cite{zhou2017unsupervised,garg2016unsupervised,flynn2016deepstereo,godard2017unsupervised} which is referred to as the photometric consistency constraint. 
Fig.~\ref{fig.3d_losses} shows the overview of 3D projections between the reference and target views while computing photometric consistency loss. 
Differ from the previous synthesis methods, we should consider the spherical nature of $360^\circ$ to project between our cubemap and 3D point cloud.

Given a pair of cubemap which have total $N$ pixels (reference $I_i$, target $I_j$) with reference depth $D_i,$ reference explainability mask $X_i$ (described in Sec.~\ref{sec.2dlosses}) and relative pose $P_{ij}$, the photometric consistency loss is defined as:
\begin{align}
	\mathcal{L}_{rec} = \frac{1}{N}\sum_i\sum^N_p X_i(p)  \cdot |I_i(p) - \hat{I}_i(p)|~,
\end{align}
which can be derived by the difference between $I_i$ and projected $\hat{I}_i$ from $I_j$
\begin{align}
\hat{I}_i  &= \xi(P_{ij} \cdot Q_i, I_j)~, \\
Q_i &= \Psi(D_i)~.
\end{align}
 
\noindent\textbf{Pose consistency.} 
Although it is straight-forward to output one global pose, we found it hard to predict pose consistently. As a result, we introduce a novel consistency loss we called "Pose Consistency."
For a cubemap, the angle difference from the front face to other faces have been given by function $\Phi$. 
For instance, the viewing angle from a front face to the right one is $90^\circ$. 
Since our network will predict the relative pose from $I_t$ to $I_{t+1}$ for each faces, the $6$ relative pose should be consistent in a cubemap coordinate. 
Hence, here we introduce Pose Consistency Loss to optimize our PoseNet.

We use $\{P_c^{F-i} = (R_c^{F-i}, 0) | i \in \mathcal{F}\}$ to represent the $6$ given constant pose from the front face to other faces in a same cube and the relative pose of each faces from $I_t$ to $I_{t+1}$ predicted by our PoseNet are denoted as $\{ P_{t, t+1}^i | i \in \mathcal{F} \}$. 
Now we can take the front face as reference view and transform all $P_{t, t+1}^i$ to the front face coordinate as ${P_{t, t+1}^{i'}}$. The rotation and translation of ${P_{t, t+1}^{i'}}$ can be derived as Eq.~\ref{eq.pose-consist}.
\begin{align}
P_{t, t+1}^{i'} &= (R_c^{i-F} \cdot R_{t, t+1}^i \cdot R_c^{F-i},~R_c^{i-F} \cdot T_{t, t+1}^i)~,
\label{eq.pose-consist}
\end{align}
where $R_{t, t+1}^i, T_{t, t+1}^i$ is the rotation and translation of $P_{t, t+1}^{i}$. Optimally, all $P_{t, t+1}^{i'}$ should be close to each other because camera poses of each faces are innately coherent. A simple case is when $i=R$  as shown in Fig.~\ref{fig.3d_losses}. We first rotate the zero pose $P_c^{F-F}$ by $90^\circ$ and multiply $P_{t,t+1}^{R}$to it to get an intermediate pose $P_{t,t+1}^{F-R}$. And then rotate this pose by $-90^\circ$ to get the final transformed pose $P_{t,t+1}^{R'}$. To use this relation as a constraint to improve our PoseNet, we can try to minimize the discrepancy of all $P_{t,t+1}^{i'}$. For this purpose, we use standard derivation as our Pose Consistency Loss as in Eq.~\ref{eq.pc}.
\begin{align}
\label{eq.pc}
\mathcal{L}_{pose} &= \sqrt{ \frac{  \sum\limits_{i \in f}{  (P_{t,t+1}^{i'} - P_{t,t+1}^*)^2 }  }{6} }~,
\end{align}
where $P_{t, t+1}^* = \frac{1}{6} \sum\limits_{i \in f}{P_{t,t+1}^{i'}}$, the mean of $6$ poses. 

\subsection{2D image-based constraints}\label{sec.2dlosses}

In addition to the 3D constraints, we further adopt several 2D constraints in our training objectives.

\noindent\textbf{Explainability mask} \label{sec.exp_mask}
To tackle the non-static objects and occlusion between views, our PoseNet outputs explainability masks in levels corresponding to depth prediction. \cite{zhou2017unsupervised} proposed a useful explainability mask learning objective

\begin{align}
    \mathcal{L}_{exp} = -\frac{1}{N}\sum_i\sum^N_p\log X_i(p) \label{eq.expl}~,
\end{align}
where $p$ denotes the pixel position and $N$ is the total amount of pixels. Eq. ~\ref{eq.expl} can be seen as a cross-entropy loss with constant ``one''  label. We found that the masks learn to interpret occlusion and high-frequency areas where photometric consistency mis-matched.

\noindent\textbf{Smoothness regularization} \label{sec.smooth}
A smoothness loss is commonly employed for regularizing the depth prediction. One type as \cite{zhou2017unsupervised,godard2017unsupervised,garg2016unsupervised} is to derive the gradients directly from spatial regions of depth.
\begin{align}
    \mathcal{L}_{sm} = \sum_i|\nabla^2 D_i| \label{eq.sm}
\end{align}

\subsection{Final model}\label{sec.final_model}

Considering all mentioned loss terms,  we train our model with the overall objective to optimize both photometric and pose constraints together with spatial smoothing and regularization:
\begin{align}
    \mathcal{L}_{all} = \mathcal{L}_{rec}+\lambda_{pose}\mathcal{L}_{pose}+\lambda_{sm}\mathcal{L}_{sm}+\lambda_{exp}\mathcal{L}_{exp}
\end{align}
\section{Dataset}\label{sec.Dat}

\begin{figure}[t]
\begin{center}
\includegraphics[width=1\linewidth]{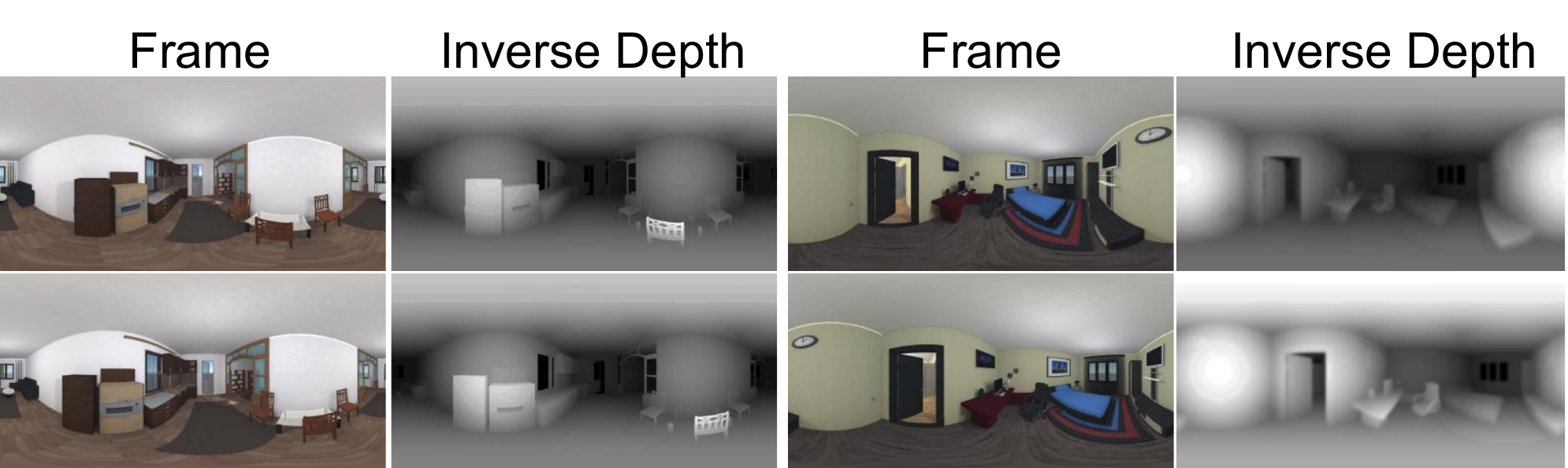}
\end{center}

\caption{
Dataset examples. We randomly pick $2$ scenes, $2$ consecutive timesteps for each, and show equirectangular RGB frames and inverse depth maps. Our dataset consists of abundant indoor scenes with high-quality depth groundtruth.
}
\label{fig.dataset}
\end{figure}



To evaluate the accuracy of our method, we collect a new dataset called --- \textbf{PanoSUNCG}. 
This dataset consists of 103 scenes with about $25k$ 360$^\circ$ images. To collect the dataset, we build up an environment with Unity using SUNCG\cite{song2016ssc}. We asked the annotators to draw $5$ trajectories in each of $103$ scenes, which should be reasonable path avoiding obstacles. Some sample snapshots are shown in Fig.~\ref{fig.dataset}.
Following a trajectory in the 3D synthetic scene, all view and depth information can be rendered into equirectangular images.
To the best of our knowledge, the proposed PanoSUNCG dataset is the first 360$^\circ$ indoor video dataset with panorama depth and camera pose groundtruth sequence. By using the synthetic environment, this dataset can be easily scaled up to have more scenes, more trajectories, or even better quality of rendering.
In our experiment, $80$ of scenes are used for training and $23$ are used for testing. 
\hu{To foster future research in 360$^\circ$ videos, we provide PanoSUNCG dataset at our \footnote{https://aliensunmin.github.io/project/360-depth/}{website}.}


\section{Experiments}\label{sec.Exp}

We conduct several experiments to evaluate our method. In Sec.~\ref{sec.impl} we describe our training details and the incremental training method, then we describe different settings and variants in Sec.~\ref{sec.baseline}. Lastly in Sec.~\ref{sec.result_eval}  we show the result of depth and camera pose estimation.

\subsection{Implementation detail} \label{sec.impl}

\textbf{Training details.}
We implemented our network using PyTorch~\cite{paszkepytorch} framework. 
The network is modified based on SfMLearner~\cite{zhou2017unsupervised} backbone. During optimization, we used Adam optimizer with ${\beta_1=0.9}$, ${\beta_2=0.999}$, initial learning rate of ${0.0006}$ and mini-batch size of $4$. For the hyperparameters of our training loss, we found that $\lambda_{pose} = 0.1$, $\lambda_{sm}=0.04$ and $\lambda_{exp} = 0.3$ balance each term in loss dropping. We train all our model 30k-50k iterations.

\noindent\textbf{Incremental training.}
Theoretically, PoseNet and DispNet can improve each other simultaneously. 
However, we found that, at the beginning of training, the two networks are unstable and will suppress each other. 
When this occurs, both depth and camera motion estimation have problematic convergence behaviors. 
To increase training stability, we first fit both networks with a relatively small dataset (4 batch size) to initialize the parameters to a stable state. 
During our training, we gradually increase the amount of our training dataset. 
When average loss is lower than an update threshold $\gamma$, we double our data and update $\gamma$ with a factor of $1.2$. 
In our experiments, we found $0.16$ as a good initial $\gamma$.

\subsection{Experimental setups} \label{sec.baseline}

In the real case when 360$^\circ$ depth is acquired, there might be an additional input: camera motion signal from the camera device. 
To verify the capacity of our method in different types of supervision signals, we set up different conditions to train our model. Also, we compare our approach with the method which directly using an equirectangular image as input to prove cubemap representation can improve 360$^\circ$ depth prediction.

\noindent\textsl{\underline{No PoseGT (fully unsupervised)}} ---
Our network can provide a reasonable camera motion estimation without any given labels. 
By applying spherical photo consistency, the self-supervised network can learn by watching 360$^{\circ}$ videos without access to camera motions.
With cube padding~\cite{Cheng_2018_CVPR} that enhances the bounding of $6$ faces, our network can learn inter cube face relationship at no cost.

\noindent\textsl{\underline{Noisy PoseGT}} ---
In the real world, estimated signals carry bias more or less. 
And small inertial sensor error can accumulate to a drastically erroneous influence along with time. 
According to \cite{phoneimu16}, a normal cell phone has about $8.1 mrad/s$ biases in gyroscope and $15mg_0$ for X- and Y-axes, $25mg_0$ for Z-axes biases in accelerometer measurements. 
We apply this constant bias error with scale on PoseGT to examine the robustness of our model. 

\noindent\textsl{\underline{PoseGT}} ---
With our dataset, we can leverage the available camera motion information as a self-supervision signal.
By exploring in a simulated environment, error-free motions can be seen as the output of optimal PoseNet, and positively improve the depth network reconstruction accuracy.
As a result, we can view this setting as the ideal situation in a real-world application.

\noindent\textsl{\underline{Equirectangular (EQUI)}} --- 
The equirectangular images instead of its cubemaps are adopted as the inputs of the network.
 
\noindent\textsl{\underline{Single pose}} --- 
To prove applying pose consistency in Sec.~\ref{sec.losses} can improve the performance of pose prediction, in the PoseNet, we concatenate all bottleneck features from the six faces and apply a 2D convolution and global average pooling along the width and height dimensions, which is similar to \cite{zhou2017unsupervised}, to make PoseNet output only one pose.


\subsection{Evaluation of Depth and Camera Motion Estimation}\label{sec.result_eval}
We conduct experiments to benchmarking our performance with variants in Sec.~\ref{sec.baseline}. 
In the following, we will visualize our depth estimation result and show quantitative results of both depth and camera motion on PanoSUNCG. In addition, to prove the potential of practical applications using our method, we conducted experiments on real-world videos.

\begin{figure}[h!]
\begin{center}
\includegraphics[width=1\linewidth]{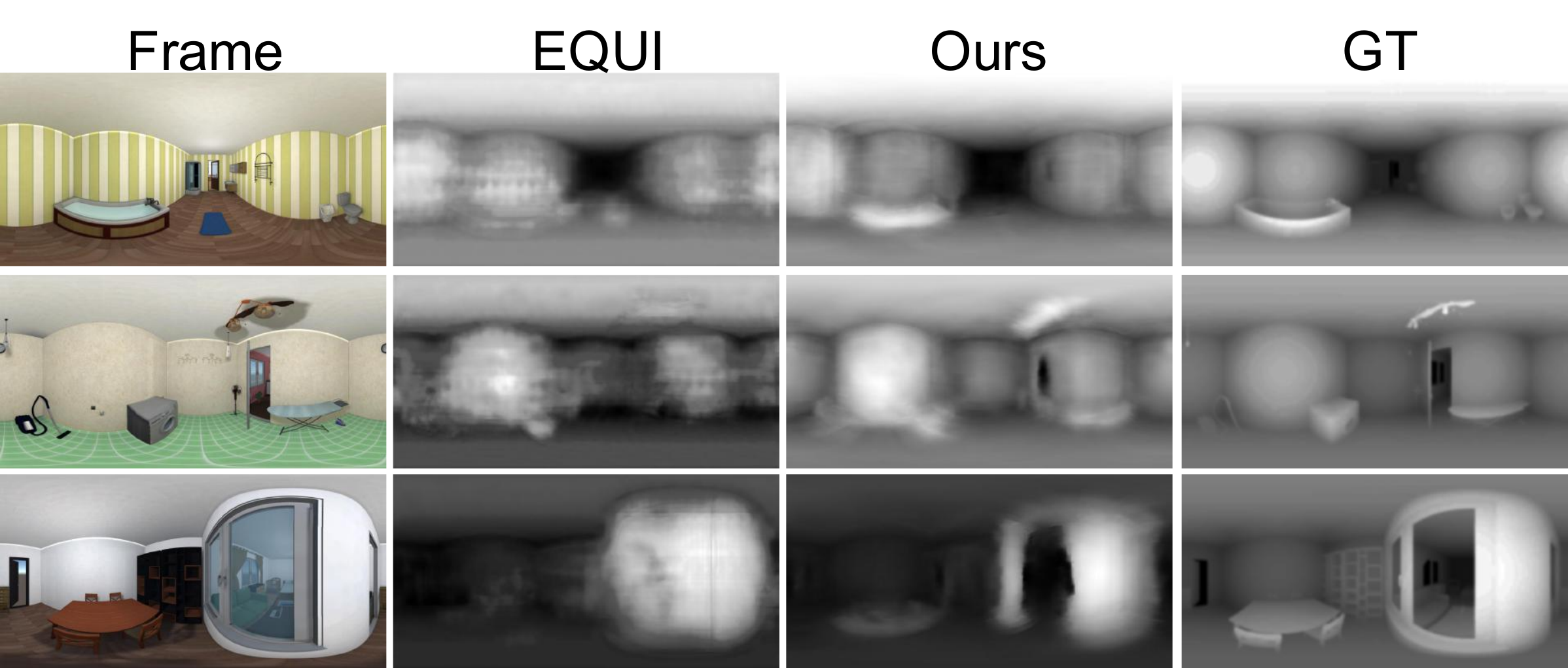}
\end{center}
\caption{
Our depth prediction in PanoSUNCG comparing to equirectangular (EQUI) baseline.
}
\label{fig.panosuncg_result2}
\end{figure}
\begin{figure}[h!]
\begin{center}
\includegraphics[width=1\linewidth]{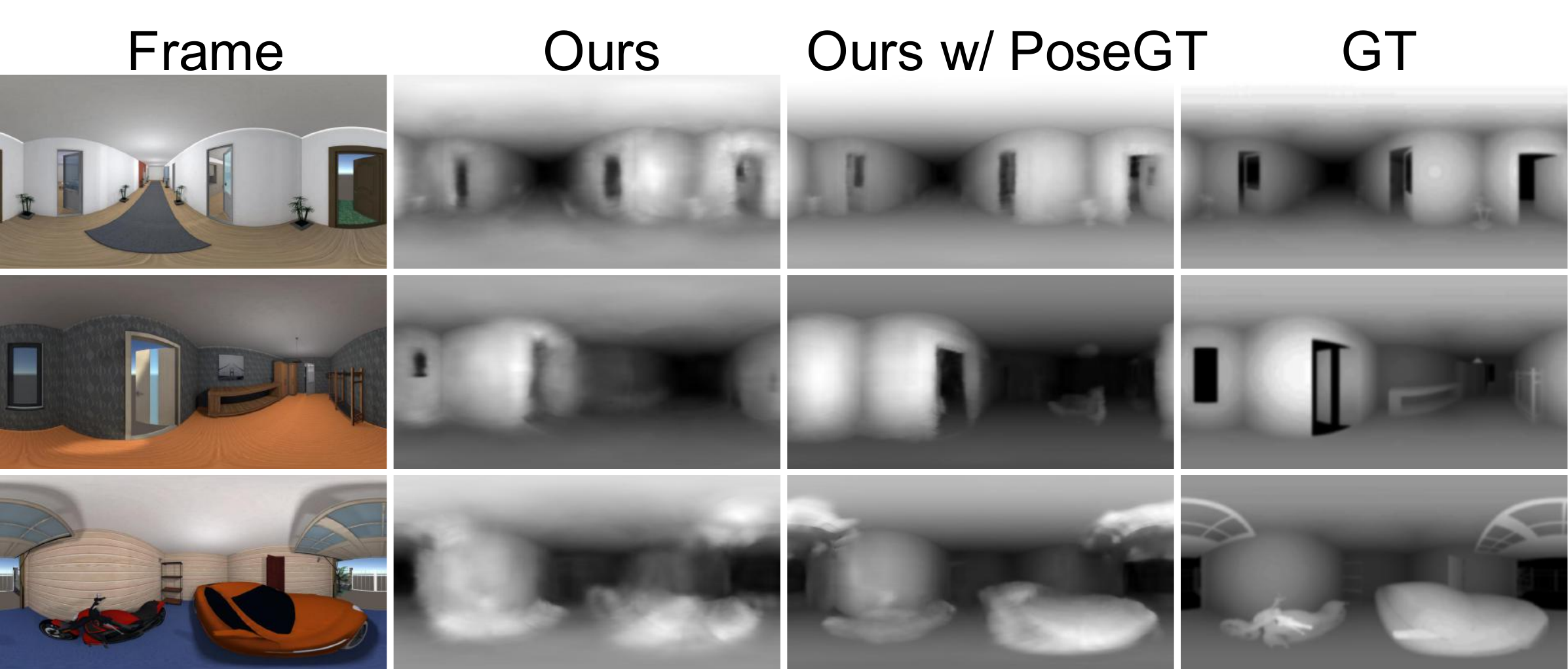}
\end{center}
\caption{
Our depth prediction in PanoSUNCG using different level of supervision.
}
\label{fig.panosuncg_result1}
\end{figure}

\noindent\textbf{Results on PanoSUNCG.}
To test the reliability of our method, we evaluate the accuracy on PanoSUNCG. As shown in Tab.~\ref{tab.depth_eval}, Fig.~\ref{fig.panosuncg_result2} and Fig.~\ref{fig.panosuncg_result1}, by considering spherical photometric consistency and pose consistency, our method can predict better depth than other baselines described in Sec.~\ref{sec.baseline}. For camera motion estimation, we use Relative Pose Error (RPE) as our evaluation metric and Table~\ref{tab.pose_eval} is our pose accuracy. The single pose baseline is using the global pose from PoseNet as described in Sec.~\ref{sec.baseline}, which is worse than our method. We also tried to train our model (both EQUI and cubemap) by the method proposed in \cite{zhou2017unsupervised}, but it ended up with bad depth and camera motion quality. It is obvious that pinhole model projection in \cite{zhou2017unsupervised} is not suitable for our 360$^\circ$ geometry.

\begin{table}[htbp]
  \centering
    \begin{tabular}{|l|c|c|c|c||c|c|c|}
    \hline
    Method & Abs Rel & Sq Rel & RMSE & $RMSE_{log}$ & $\delta < 1.25$ & $\delta < 1.25^2$ & $\delta < 1.25^3$ \bigstrut\\
    \hline \hline
    \textbf{Ours (Full Model)} & \textbf{0.337} & \textbf{4.986}	& \textbf{8.589} & 0.611 & \textbf{0.647} & \textbf{0.829} & \textbf{0.899} \bigstrut[t]\\
    Ours w/o $\mathcal{L}_{pose}$ & 0.418 & 7.113 & 9.916 & 0.698 & 0.580 & 0.790 & 0.876 \\
    EQUI & 0.395 & 7.279 & 9.405 & \textbf{0.493} & 0.623 & 0.803 & 0.880 \bigstrut[t]\\
    \hline \hline
    Ours w/ PoseGT & 0.254 & 3.554 & 7.126 & 0.513 & 0.752 & 0.880 & 0.927 \bigstrut[t]\\
    Ours w/ Noisy PoseGT & 0.283 & 4.200 & 7.636 & 0.498 & 0.722 & 0.867 & 0.919 \bigstrut[b]\\
    \hline
    \end{tabular}%
   \caption{
Depth prediction result on PanoSUNCG. Following metrics used in \cite{zhou2017unsupervised,garg2016unsupervised,godard2017unsupervised}, we compare our method with ablation settings, baselines, and different level of supervision. Note that the numbers in the first 4 columns are the lower the better, and the last 3 columns higher the better.
   }
  \label{tab.depth_eval}%
\end{table}%

\begin{table*}[t]
\centering
\begin{tabular}{|l|c|c|} \hline
Method & RPE-R & RPE-T  \\ \hline \hline

\textbf{Ours (Full Model)} & \textbf{6.98} &  \textbf{0.025} \\ 
Ours w/o $\mathcal{L}_{pose}$  & 7.37	&  0.039\\ 
Single pose  &  7.49 & 0.026 \\   
EQUI & 7.06 & 0.025 \\ 
\hline
\end{tabular}
\begin{minipage}{0.5\textwidth}
\caption{RPE of camera motion estimation.   
RPE-R and RPE-T is rotation (in degree) and translation error, respectively. Our full method predicts better camera motion than other baselines.
}
\label{tab.pose_eval}
\end{minipage}
\end{table*}

\noindent\textbf{Results on real-world videos.}
To demonstrate the usability of our method on real-world cases, we record our own 360$^\circ$ videos by RICOH THETA V. 
We extract the videos at $5$ fps and finetune the PanoSUNCG-pretrained model on a train split.
Fig.~\ref{fig.realworld_result} shows the qualitative result of image and predicted depth pairs by our proposed self-supervised method.

For clearer visualization, we also plot the point cloud from the predicted depth. The results show that our proposed method produces promising results in real-world scenarios.

\subsection{Time Evaluation}\label{sec.times}
\begin{figure}[t!]
\begin{center}
\includegraphics[width=\linewidth]{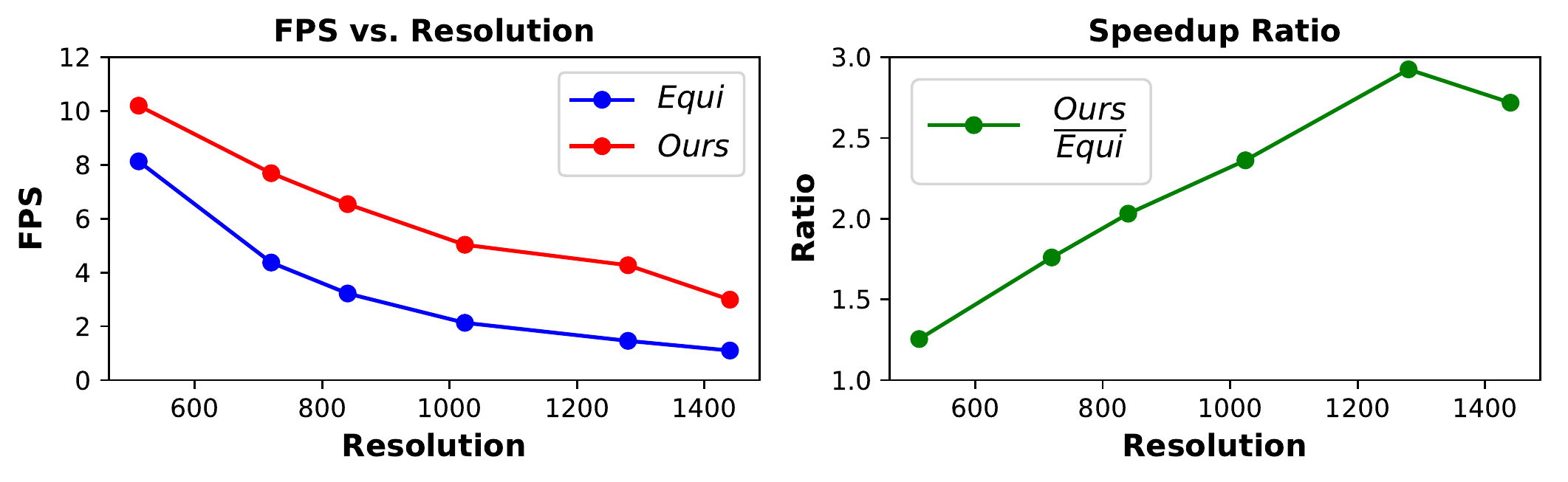}
\end{center}
\caption{
Frame per second measurement of EQUI and Our method. Our method can perform faster inference speed than EQUI baseline. The speedup ratio increase more than 200\% given higher input resolution.
}
\label{fig.speed}
\end{figure}

\begin{figure}[h!]
\begin{center}
\includegraphics[width=1\linewidth]{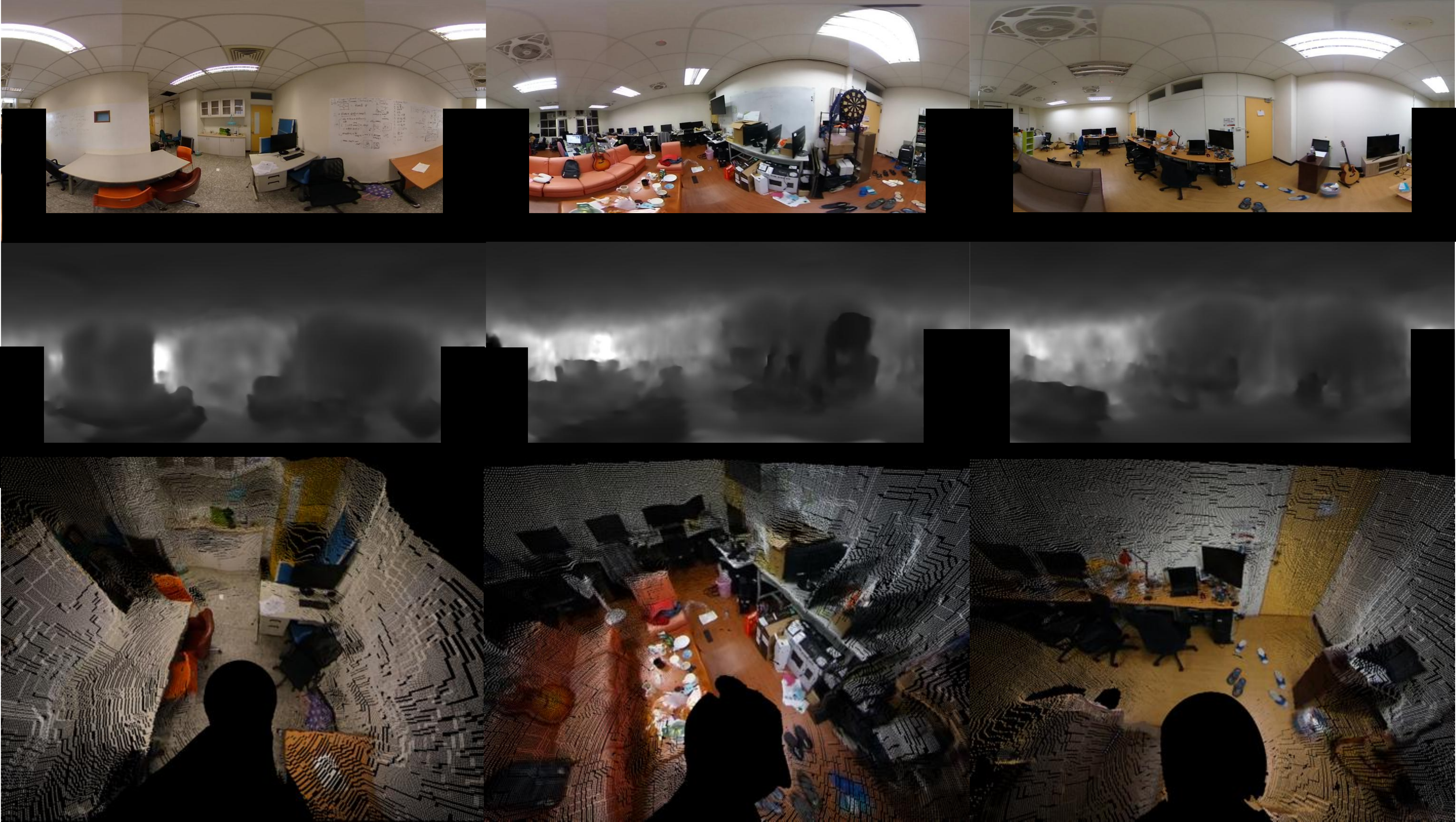}
\end{center}
\caption{
Our depth prediction in real-world videos (image, depth and point cloud in different scenarios).
}
\label{fig.realworld_result}
\end{figure}

For equirectangular images, there exists a considerable distortion in the field near two polars, consisting of a lot of unnecessary pixels. Fortunately, for a cubemap, there are no redundant pixels because it consists of 6 perspective images. As a result, the most significant benefit of using cubemap is the huge reduction in computing time.
In Fig.~\ref{fig.speed}, we plot frame rate and their speed up ratio as a function of image resolution. We denote the height of the equirectangular image as resolution. 
Our method, therefore, takes cubemap with width equals to $resolution/2$. 
The time computation of our method includes both converting equirectangular image to cubemap and converting cubemap depth back to equirectangular depth, which is a fair comparison. 
As we could see, the efficiency gap is getting larger as we have larger image resolution. 
Therefore, using cubemap can significantly reduce the computation cost, which means our method is much more suitable to be productized in an autonomous system or robotics application.
Similar comparisons to other baseline methods are shown in the technical report \cite{WangHuACCVsupp18}.

\section{Conclusion}\label{sec.Con}
We have presented a self-supervised method for depth and camera motion estimation in 360$^{\circ}$ videos. To overcome the projection distortion of equirectangular, our entire model processes data in cubemap, incorporating Cube Padding~\cite{Cheng_2018_CVPR} to join every cube faces spatially. Furthermore, we proposed two novel self-training objectives tailored for geometry in 360$^{\circ}$ videos. The spherical photometric consistency loss is to minimize the difference between warped spherical images; the camera pose consistency loss is to optimize the rotation and translation difference between cube faces. The depth and camera pose estimated by our method outperform baseline methods in both quality and inference speed. Our work makes a notable step toward 3D reasoning in 360$^\circ$ videos which can be used in autonomous systems with omnidirectional perception or VR in the future.
\section{Acknowledgements}\label{sec.Ack} 
We thank MOST-107-2634-F-007-007, MOST-107-2218-E-007-047, MOST-106-2221-E-007-107 and MEDIATEK for their support.

\clearpage

\end{document}